\documentclass{article}

     \PassOptionsToPackage{numbers, compress}{natbib}
     \usepackage[final]{neurips_2021}

\usepackage[utf8]{inputenc} % allow utf-8 input
\usepackage[T1]{fontenc}    % use 8-bit T1 fonts
\usepackage{hyperref}       % hyperlinks
\usepackage{url}            % simple URL typesetting
\usepackage{booktabs}       % professional-quality tables
\usepackage{amsfonts}       % blackboard math symbols
\usepackage{nicefrac}       % compact symbols for 1/2, etc.
\usepackage{microtype}      % microtypography
\usepackage{graphicx}
\usepackage{subfigure}
\usepackage{amsmath}
\usepackage{pdfpages}
\usepackage{multirow}
\usepackage{pgfplots}
\pgfplotsset{compat=1.14}
\usetikzlibrary{calc}
\usetikzlibrary{positioning}
\usepackage{comment}

\usepackage{xcolor}

\usepackage{algorithm}  
\usepackage{algpseudocode}  
\usepackage{amsmath}  
\usepackage{bm}
\title{ML4CO-KIDA: Knowledge Inheritance in Dataset Aggregation}
\definecolor{MyRed}{rgb}{0.8,0.2,0}
\definecolor{MyBlue}{rgb}{0,0,1.0}

\author{%
  Zixuan Cao \\
  Peking University\\
  Megvii Research\\
  \texttt{caozixuan.percy@stu.pku.edu.cn} \\
  
  \And
  Yang Xu \\
  Peking University \\
  Megvii Research \\
  \texttt{1800010740@pku.edu.cn} \\
  
  \And
  Zhewei Huang \\
  Megvii Research\\
  \texttt{huangzhewei@megvii.com} \\
  
  \And
  Shuchang Zhou \\
  Megvii Research\\
  \texttt{zsc@megvii.com} \\
  }
\begin{document}

\maketitle

\begin{abstract}
The Machine Learning for Combinatorial Optimization (ML4CO) NeurIPS 2021 competition aims to improve state-of-the-art combinatorial optimization solvers by replacing key heuristic components with machine learning models. On the dual task, we design models to make branching decisions to promote the dual bound increase faster. We propose a knowledge inheritance method to generalize knowledge of different models from the dataset aggregation process, named KIDA. Our improvement overcomes some defects of the baseline graph-neural-networks-based methods. Further, we won the $1$\textsuperscript{st} Place on the dual task. We hope this report can provide useful experience for developers and researchers. The code is available at \url{https://github.com/megvii-research/NeurIPS2021-ML4CO-KIDA}.
\end{abstract}
% \usepackage{multirow}
% \usepackage{multirow}
% \usepackage{multirow}

In the dual task of ML4CO NeurIPS $2021$ competition, as shown in Table~\ref{table:rank}, our team (Nuri) ranks $1$\textsuperscript{st}, $6$\textsuperscript{th}, and $1$\textsuperscript{st} in three benchmarks of Balanced Item Placement, Workload Apportionment, and Anonymous Problem separately. Our report is organized as follows: Section~\ref{Pre} introduces the basic information of ML4CO; Section~\ref{issue} introduces issues we observe when following the baseline method; Section~\ref{method} explains in detail the main method we develop to improve the quality of branching decisions further; Section~\ref{exp} shows the performance of different models in each benchmark; Section~\ref{discuss} discusses the failure of Strong Branching~\cite{achterberg2005branching} which is regarded as expert knowledge; We conclude our report in Section~\ref{conclude}.       
% \usepackage{multirow}
\begin{table}[hb]
\centering
\caption{\textbf{Leaderboard of the Dual Task in ML4CO NeurIPS 2021 competition}}
\label{table:rank}
\resizebox{1\textwidth}{!}{
\begin{tabular}{lclclcll}
\toprule
\multirow{2}{*}{Team} & \multicolumn{2}{c}{Item Placement}   & \multicolumn{2}{c}{Load Balancing}     & \multicolumn{2}{c}{Anonymous}            & \multirow{2}{*}{Score~$\downarrow$} \\ \cline{2-7}
                      & \multicolumn{2}{c}{Cum. Reward~$\uparrow$}      & \multicolumn{2}{c}{Cum. Reward~$\uparrow$}        & \multicolumn{2}{c}{Cum. Reward~$\uparrow$}          &                        \\ \midrule
\textbf{Nuri~(Ours)}         & \multicolumn{2}{c}{\textbf{6684.00}} & \multicolumn{2}{c}{630787.18}          & \multicolumn{2}{c}{\textbf{27810782.42}} & $\bm{6}~(1\times6\times1)$            \\
EI-OROAS              & \multicolumn{2}{c}{6670.30}          & \multicolumn{2}{c}{\textbf{631744.31}} & \multicolumn{2}{c}{27158442.74}          & $8~(2\times1\times4)$                      \\
EFPP                  & \multicolumn{2}{c}{6487.53}          & \multicolumn{2}{c}{631365.02}          & \multicolumn{2}{c}{26340264.47}          & $117~(3\times3\times13)$                    \\
KAIST\_OSI            & \multicolumn{2}{c}{6196.56}          & \multicolumn{2}{c}{631410.58}          & \multicolumn{2}{c}{26626410.86}          & $126~(7\times2\times9)$                    \\
qqy                   & \multicolumn{2}{c}{6377.23}          & \multicolumn{2}{c}{630557.31}          & \multicolumn{2}{c}{27221499.03}          & $132~(6\times11\times2)$                    \\ \bottomrule
\end{tabular}
}
\end{table}
\section{Preliminary}
\label{Pre}
We will introduce the background knowledge needed in the dual task and the official baseline based on graph neural network~(GNN)~\cite{scarselli2008graph, kipf2016semi}.

\subsection{Dual Task}

\begin{figure}[ht]
\centering
\includegraphics[width=0.7\linewidth]{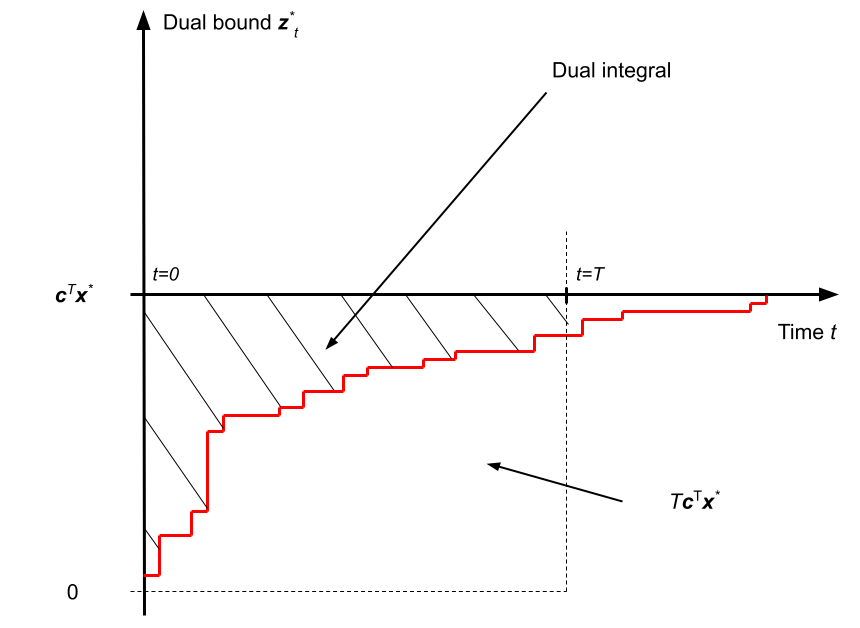}
\caption{\textbf{Evaluation Metric of the Dual Task}. Our goal is to minimize the dual integral over time}
\label{fig:dual}
\end{figure}

For mixed-integer linear programs (MILPs), a well-known solving method is branch-and-bound (B\&B)~\cite{land2010automatic}. One of the key issues of B\&B is how to select variables for \textbf{Branching}. In ML4CO competition, the goal of the dual task is to promote the dual bound to increase faster by selecting proper branching variables. The final metric is the dual integral over time, shown in Figure~\ref{fig:dual}.

\subsection{Branching Algorithms}

Although making branching decisions has received little theoretical understanding to this day~\cite{lodi2017learning}, there are many heuristic algorithms. Strong Branching involves testing which of the candidate variable gives the best improvement to the objective function before actually branching on them. Strong Branching is a high-quality but computationally expensive method. In practice, modern B\&B solvers rely on hybrid-branching~\cite{linderoth1999computational} and reliability branching~\cite{achterberg2005branching}, which only use Strong Branching at certain nodes of the search tree and use pseudo-cost branching~\cite{benichou1971experiments} at other nodes. 

\subsection{Official Baseline}

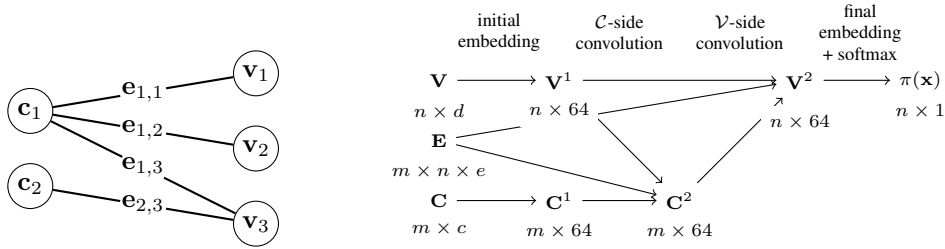
\begin{figure}[t]
\centering
\hspace{\fill}
\begin{tikzpicture}[scale=1]
    \tikzstyle{every node}=[draw,circle,minimum size=17pt,inner sep=0pt]
    \tikzstyle{annot} = [text width=6em, text centered]

    \foreach \y in {1, 2, 3}{
        \node (v\y) at (0,-\y) {$\mathbf{v}_\y$};
    }

    \foreach \y in {1, 2}{
        \path[yshift=-0.5cm, xshift=-3cm]
            node (c\y) at (0,-\y) {$\mathbf{c}_\y$};
    }

    \path (c1) edge[-, thick] node[fill=white, draw=none] {$\mathbf{e}_{1, 1}$} (v1);
    \path (c1) edge[-, thick] node[fill=white, draw=none] {$\mathbf{e}_{1, 3}$} (v3);
    \path (c1) edge[-, thick] node[fill=white, draw=none] {$\mathbf{e}_{1, 2}$} (v2);
    \path (c2) edge[-, thick] node[fill=white, draw=none] {$\mathbf{e}_{2, 3}$} (v3);

\end{tikzpicture}
\hspace{\fill}
\hspace{\fill}
\begin{tikzpicture}[scale=0.8]
\tikzset{font=\scriptsize}

\node (C0) at (0, 0) {$\mathbf{C}$};
\node (E0) at (0, 1) {$\mathbf{E}$};
\node (V0) at (0, 2) {$\mathbf{V}$};

\node (C1) at (2, 0) {$\mathbf{C}^1$};
% \node (E1) at (2, 1) {$\mathbf{E}^1$};
\node (V1) at (2, 2) {$\mathbf{V}^1$};

\node (C2) at (4, 0) {$\mathbf{C}^2$};

\node (V2) at (6, 2) {$\mathbf{V}^2$};

\node (V3) at (8, 2) {$\pi(\mathbf{x})$};

\path[draw,->] (C0) -- (C1);
% \path[draw,->] (E0) -- (E1);
\path[draw,->] (V0) -- (V1);

\path[draw,->] (C1) -- (C2);
\path[draw,->] (E0) -- (C2);
\path[draw,->] (V1) -- (C2);

\path[draw,->] (C2) -- (V2);
\path[draw,->] (E0) -- (V2);
\path[draw,->] (V1) -- (V2);

\path[draw,->] (V2) -- (V3);

\node[fill=white, below=-0.01 of V0] {$n \times d$};
\node[fill=white, below=-0.01 of E0] {$m \times n \times e$};
\node[fill=white, below=-0.05 of C0] {$m \times c$};
\node[fill=white, below=-0.05 of V1] {$n \times 64$};
% \node[fill=white, below=-0.05 of E1] {$m \times n \times 64$};
\node[fill=white, below=-0.05 of C1] {$m \times 64$};
\node[fill=white, below=-0.05 of C2] {$m \times 64$};
\node[fill=white, below=+0.10 of V2] {$n \times 64$};
\node[fill=white, below=-0.05 of V3] {$n \times 1$};

\node[align=center] at (1, 2.8) {initial\\embedding};
\node[align=center] at (3, 2.8) {$\mathcal{C}$-side\\convolution};
\node[align=center] at (5, 2.8) {$\mathcal{V}$-side\\convolution};
\node[align=center] at (7, 2.8) {final\\embedding\\+ softmax};

\end{tikzpicture}
\hspace{\fill}
\caption{Copy from ~\cite{gasse2019exact}. Left: the bipartite state representation $\mathbf{s}_t = (\mathcal{G}, \mathbf{C}, \mathbf{E}, \mathbf{V})$ with $n=3$ variables and $m=2$ constraints. Right: the bipartite GCNN architecture for parametrizing baseline policy $\pi_\theta(\mathbf{a} \mid \mathbf{s}_t)$}
\label{gnn}
\end{figure}

The baseline method~\cite{gasse2019exact} regards the decision of Strong Branching as expert knowledge and train models by imitating the expert knowledge. Specifically, we convert the combinatorial optimization problems to bipartite graphs, illustrated in Figure~\ref{gnn}. Then we use the GNN to obtain the embedding of each variable. These embeddings are used to generate branching strategies for the current state. We train this GNN by minimizing the cross-entropy loss between its output and the expert decision.  

\section{Bad Generalization Issue}
\label{issue}
Following the method in \cite{gasse2019exact}, we summarize two issues that may reduce the final performance.

\begin{figure}[h]
  \centering
  \includegraphics[scale=0.6]{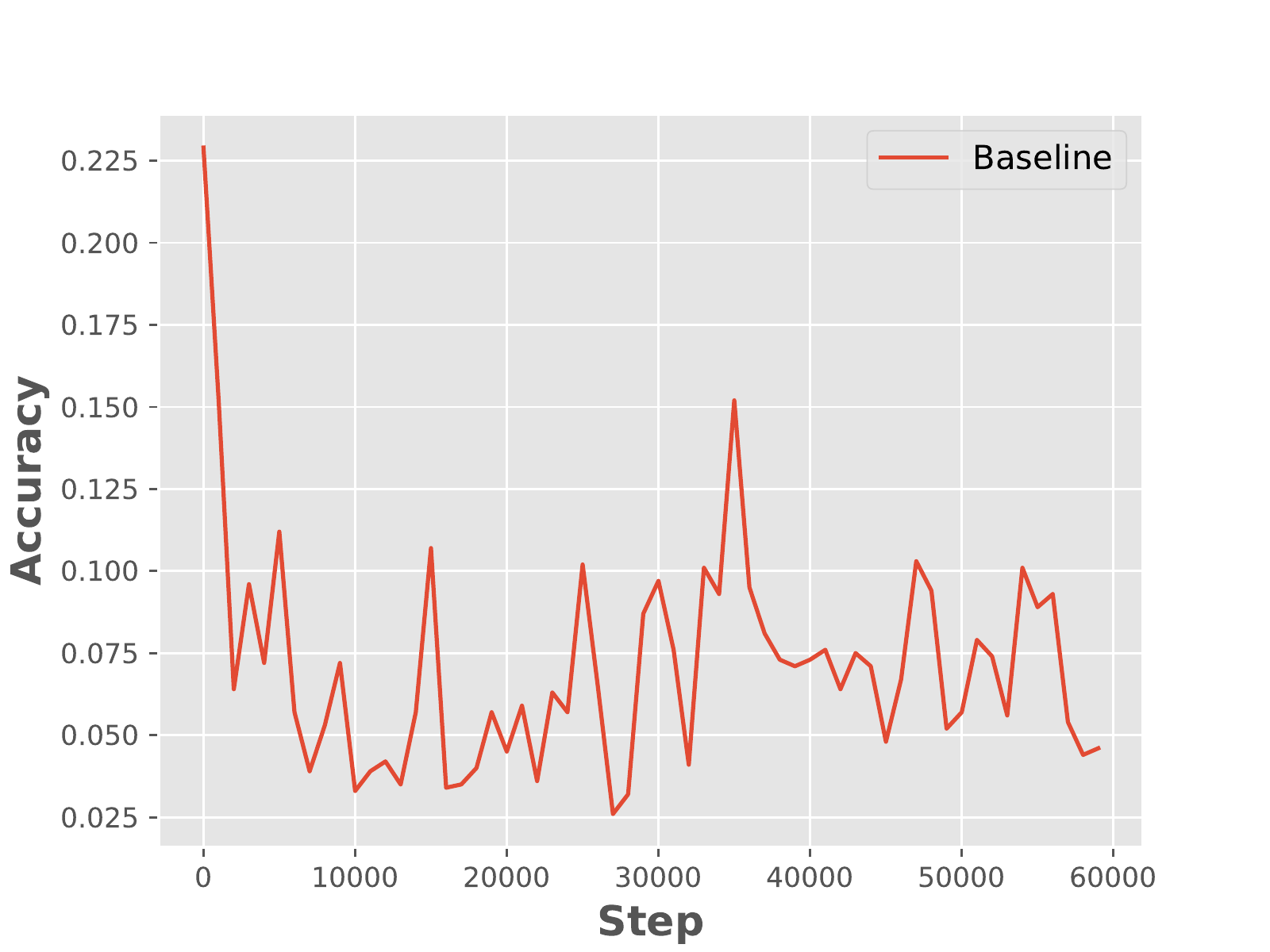}
  \caption{\textbf{Accuracy when Interacting with SCIP Solver.} During the deployment phase, the accuracy drops a lot}
  \label{fig:acc_base}
\end{figure}

First, there is a major gap between offline training and the online deployment phase. In offline training, the data is collected beforehand. The GNN model achieves an accuracy of around $0.80$~(validation). However, the accuracy shows a sharp decline when deploying a trained model. As shown in Figure \ref{fig:acc_base}, when making branching decisions for a random online instance, the averaged accuracy of different steps is about $0.1$.

Second, the higher accuracy does not necessarily lead to a higher reward. As shown in Table \ref{table:epochs}, comparing models in different epochs, the accuracy is not strongly related to the final reward.

\begin{table}[!htbp]
\centering  
\caption{\textbf{Performance Comparison of Different Epochs}. The imitation accuracy is not directly related to model performance}
\label{table:epochs}
\begin{tabular}{ccccc}
\toprule
Epoch    & Top 1 Acc.     & Top 3 Acc.     & Top 5 Acc. & Cum. Reward     \\ \midrule
1 & 0.780 & 0.932 & 0.971           & 5202.6          \\
5 & 0.803          & 0.948          &  0.981          & \textbf{5545.7}    \\ 
10 & 0.808          &  0.946         & 0.978       & 5131.9          \\ 
20 & \textbf{0.810}         & \textbf{0.947}          & \textbf{0.979}       & 5038.8         \\ \bottomrule
\end{tabular}
\end{table}
\section{Method}
\label{method}
\subsection{DAgger}

The major performance gap between training and deployment can be explained by the defects of Behavior Cloning~\cite{bain1995framework}, which forces agents to learn expert behavior. There is a common problem that the distribution of collected data and the data the model encounters in the real scene may be inconsistent~\cite{daume2009search, ross2010efficient}. For example, a driving-related dataset is rich in driver's operation data during normal driving. But when the car is about to hit the wall, there may be a lack of data to deal with the contingency. To make a more completely collected dataset, we need to constantly explore the unseen data and label these data with expert knowledge. 

Dataset Aggregation (DAgger)~\cite{ross2011reduction} is such a method to reduce the gap between training and depolyment. Based on DAgger, we collect data iteratively and train new models constantly. Details are described in Algorithm~\ref{algo:1}. 
\begin{algorithm}[h]  
\caption{DAgger in Dual Task} 
\label{algo:1}
    \begin{algorithmic}[1]  
    \State Initialize a random model $\pi_0$ and an empty dataset $D$
    \For{each $i \in [1, 50]$}
        \State Interact with the solver with $0.95$ probability of using model $\pi_{i-1}$ and 0.05 probability of using Strong Branching 
        \State Collect data obtained by Strong Branching as $D_{i}$
        \State $D = D \cup D_{i}$
        \If{$i~mod~10==0$}
            \State Train $\pi_i$ with $100$ epochs
        \Else
            \State Train $\pi_i$ with $10$ epochs
        \EndIf
    \EndFor
    \label{code:recentEnd}  
  \end{algorithmic}  
\end{algorithm}  
\subsection{KIDA}

\begin{figure}[ht]
\centering
	\includegraphics[width=0.7\linewidth]{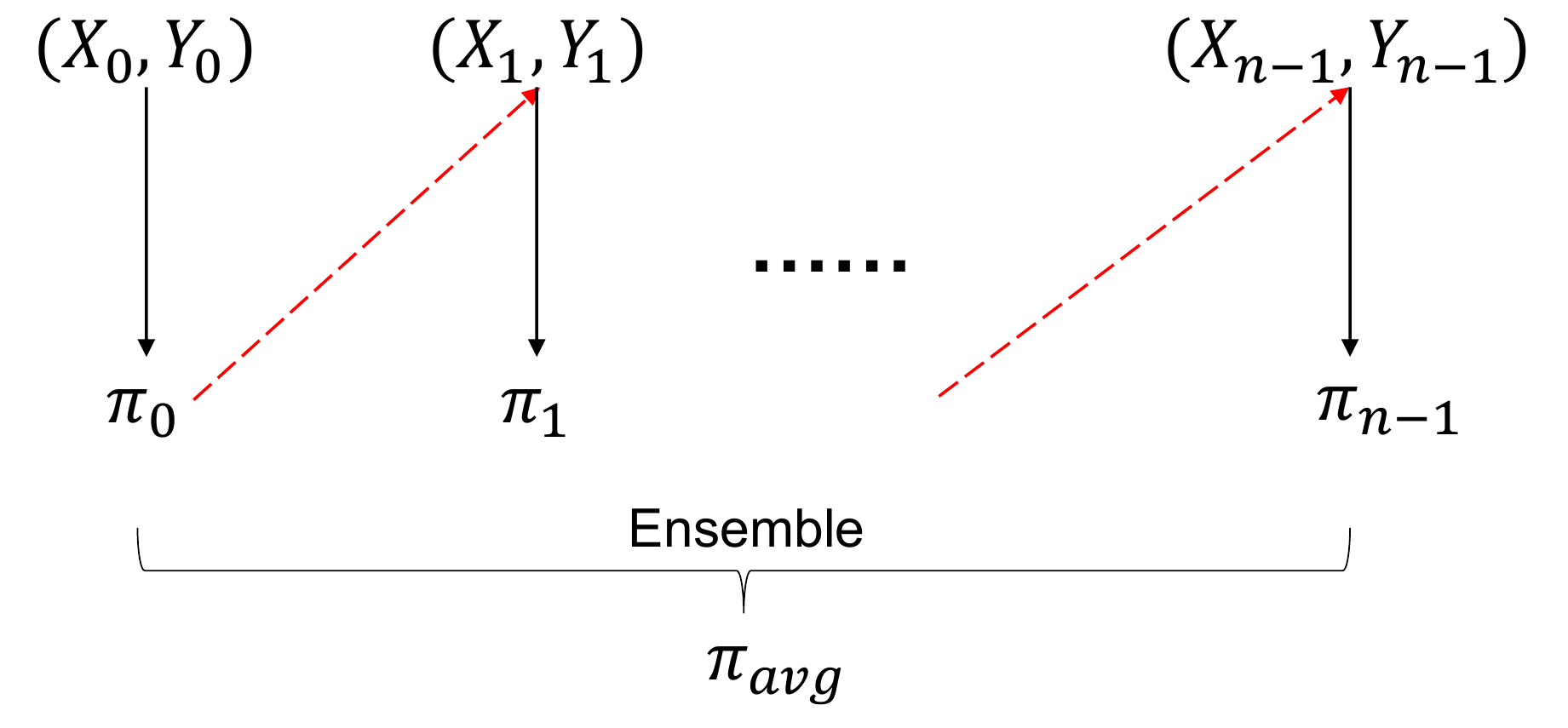}
		\label{fig:framework}
	\caption{\textbf{Framework of KIDA}. We train each model based on the data generation using the last model. Finally, we average the parameter of trained models}
\end{figure}

As shown in Section~\ref{issue}, in the Item Placement benchmark, higher accuracy does not lead to higher rewards. Experiments show that the performance of different models obtained in DAgger is not stable. Therefore, we use model ensemble~\cite{dietterich2000ensemble, kidzinski2018learning} to improve the performance of our model further. A noticeable difficulty is that the practicality of the model is time-sensitive. Averaging the output of different models is time-consuming. We consider averaging the weights of different models~\cite{tarvainen2017mean}. Here, we define Knowledge Inheritance in Dataset Aggregation (KIDA) as building a new model $\pi_{avg}$ by averaging the weights of trained models during the dataset aggregation process. The framework is shown in Figure~\ref{fig:framework}. The pipeline of KIDA is similar to Born-Again Neural Networks~\cite{furlanello2018born}. In KIDA, the training of the current model depends on the generation using the last model. The final model is the ensemble of trained models.

Formally, for models obtained from dataset aggregation with parameters $(\theta_{0}, \theta_{1}, ..., \theta_{n-1})$, the parameters of $\pi_{avg}$ are obtained by:
$\theta_{avg}=\sum_{i=0}^{n-1}\theta_{i}/n.$

\begin{figure}[ht]
\centering
	\begin{minipage}[t]{1\textwidth}
		\centering
		\includegraphics[scale=0.55]{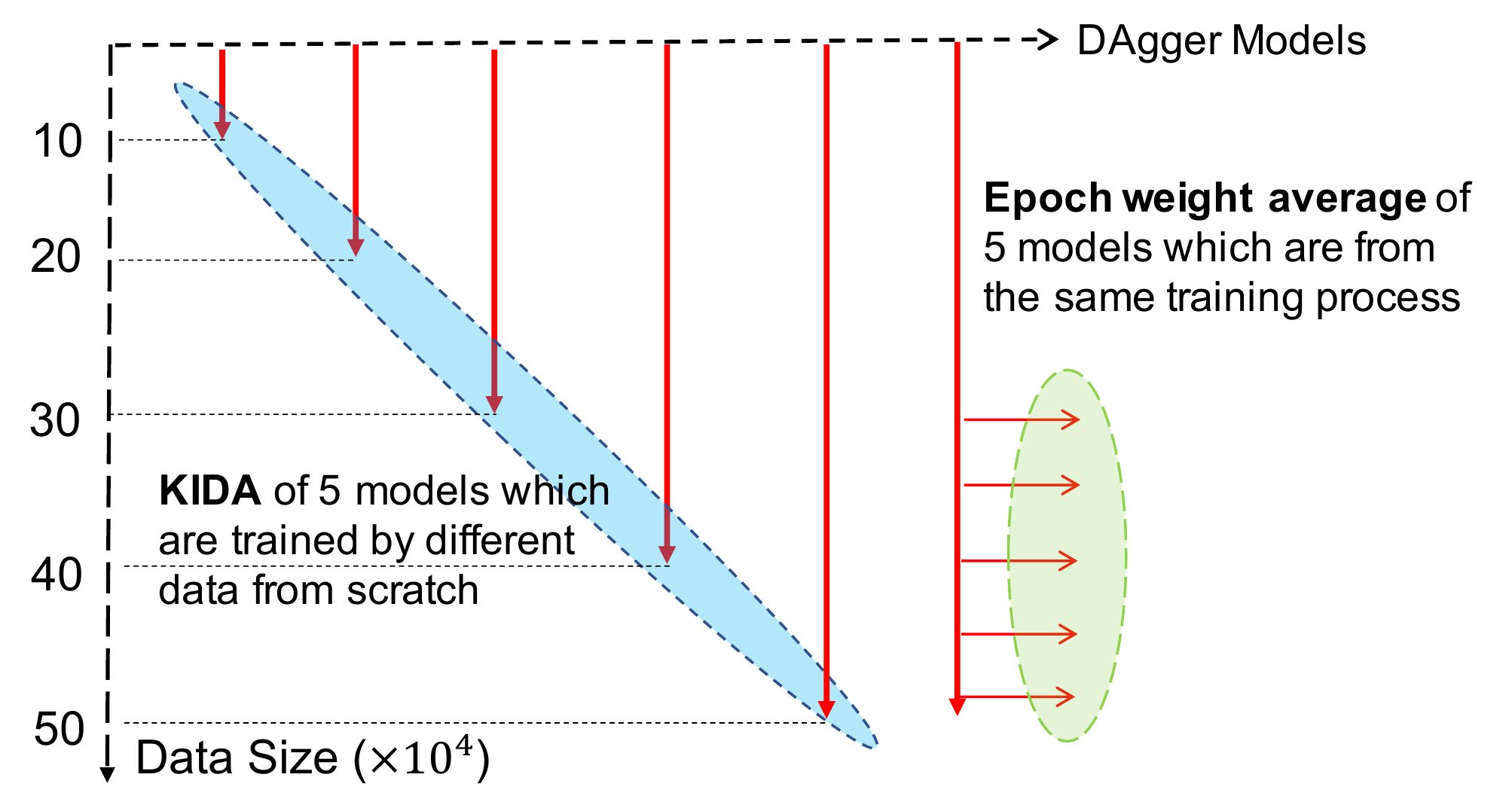}
		\label{fig:kida_ewa}
	\end{minipage}
	\caption{\textbf{KIDA vs. Epoch Weight Average}}
	\vspace{-2pt}
\end{figure}

Figure~\ref{fig:kida_ewa} shows the difference between KIDA and the popular epoch weight average method. KIDA focuses on models trained by different data from different iteration rounds of DAgger. Although the training data of models are different, these data are related. Common epoch weight average such as Snapshot ensembles~\cite{huang2017snapshot} and SWA~\cite{izmailov2018averaging} focuses on models that are obtained in the same training process of different epochs. 
%\subsection{Tricks}
%According to the specific characteristics of this benchmark, some tricks %are applied to further improve the performance of our model. 
%\begin{itemize}
%    \item Strong Branching is time-consuming in this benchmark. %Therefore, we use a longer time limit (from 15 minutes to 20 minutes) to %collect expert data.
%    \item The convolution way of \cite{gasse2019exact} is modified. For %the bipartite graph obtained from a certain CO problem, suppose the %embeddings of the node from the left part, the node from the right part, %and the edge are denoted as $L, R$, and $E$. We modify the convolution %layer from summation $R=f(g(L+R+E), R)$ to concatenation $R=f(g(concat(L, %R, E)), R)$.
%    \item Dropout layers are added at the embedding layers and final %output layer.
%\end{itemize}
\section{Experiment}
\label{exp}
%补充一下介绍
In ML4CO competition, there are three benchmarks, including Balanced Item Placement, Workload Apportionment, and Anonymous Problem \footnote{Details about each benchmark can refer to https://www.ecole.ai/2021/ml4co-competition/} which need to be evaluated separately. In this section, we discuss the performance of different models in each benchmark. 
\subsection{Anonymous Problem}

\begin{figure}[!htbp]
  \centering
  \includegraphics[scale=0.40]{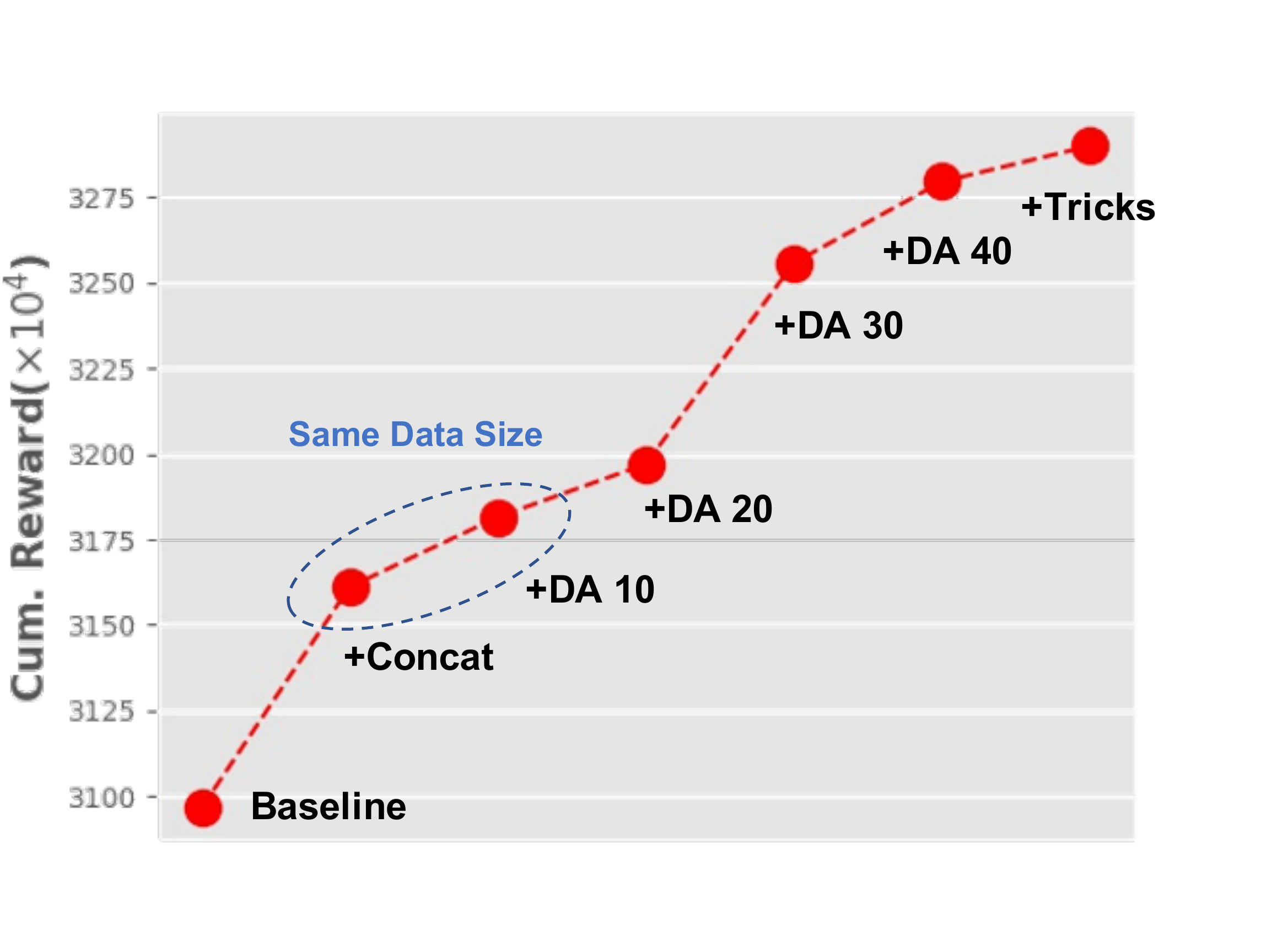}
  \caption{\textbf{The Cumulative Reward of Anonymous in Validation Dataset}. The performance of DAgger models is related to the number of iteration rounds}
   \label{fig:1}
\end{figure}

We show the performance of models with different settings in the validation set in Figure~\ref{fig:1}. With the same data size, the model using DAgger performs better. With the increase of data size, the performance increases. The model we finally submitted is a new DAgger model with Dropout~\cite{srivastava2014dropout} layers and a longer solver time when collecting data.

\subsection{Balanced Item Placement}

\begin{figure}[!htbp]
  \centering
  \includegraphics[scale=0.4]{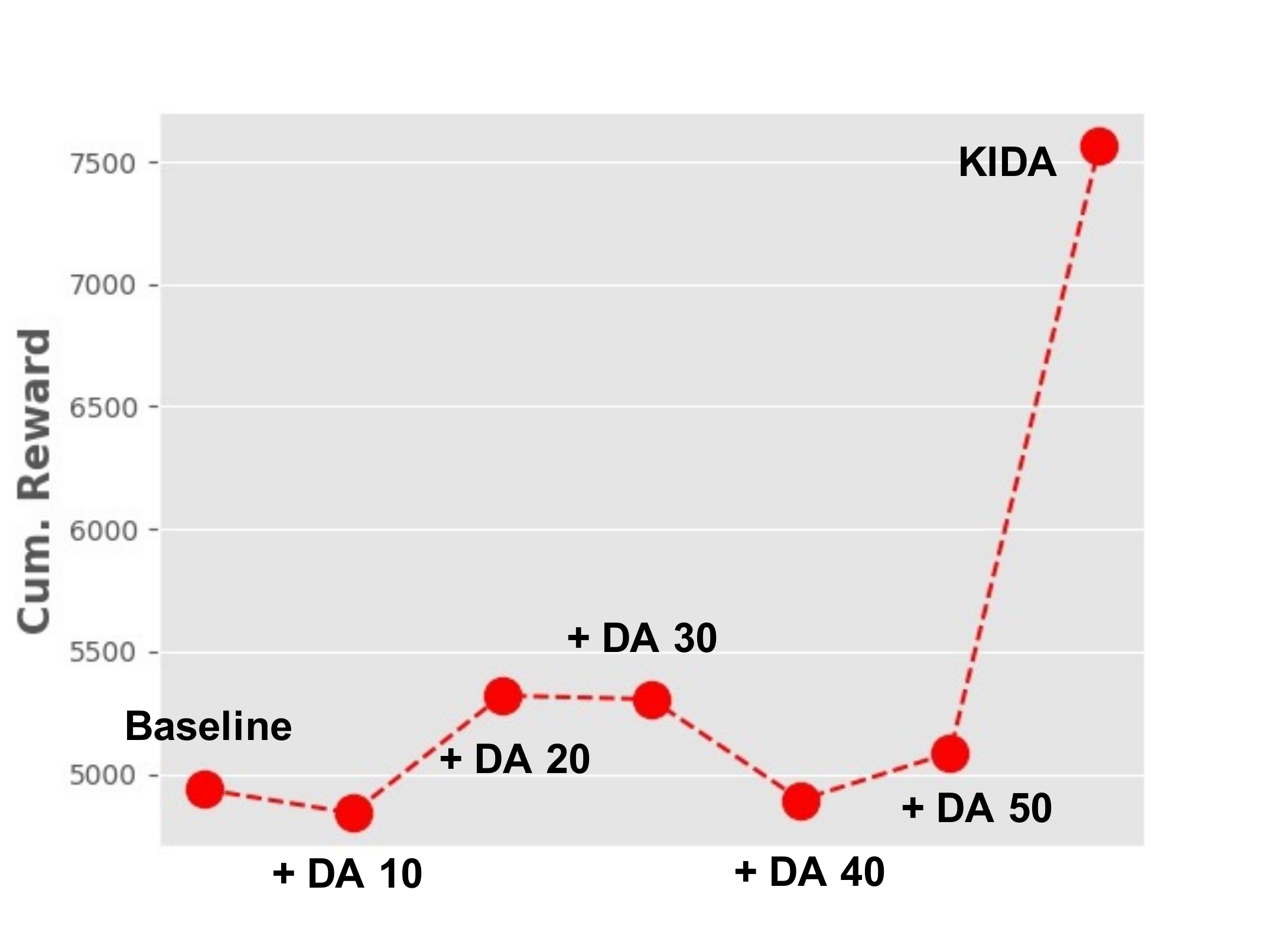}
  \caption{\textbf{The Cumulative Reward of Item Placement in Validation set}. KIDA has significant advantages over just applying DAgger}
   \label{fig:kida}
\end{figure}

\noindent\textbf{Overall Performance}. The performance of DAgger at different iteration rounds is shown in Figure~\ref{fig:kida}. Unlike in the Anonymous benchmark, the performance of DAgger is unstable. But when we apply KIDA in this benchmark, the performance has significantly improved.  

\begin{figure}[h]
  \centering
  \includegraphics[scale=0.6]{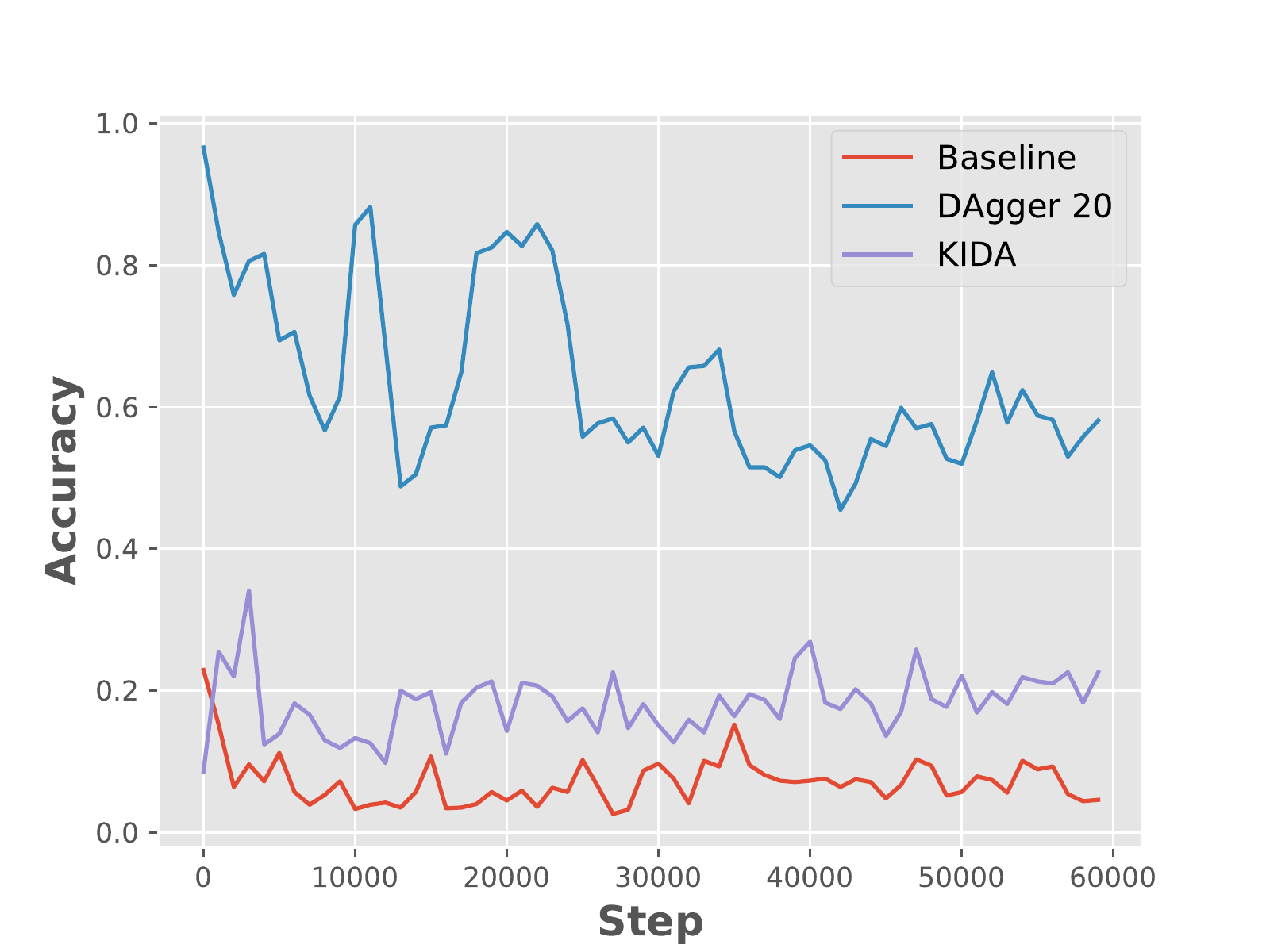}
  \caption{\textbf{Accuracy when Interacting with the Solver}. Overall accuracy of DAgger 20, KIDA and the baseline are 0.635, 0.182, 0.071 respectively.}
   \label{fig:acc}
\end{figure}

\noindent\textbf{Detailed Evaluation.} To further explore our method, we randomly process a problem from the validation set using Strong Branching and compare the output of different models with Strong Branching labels. The accuracy of different models is shown in Figure \ref{fig:acc}. DAgger model still has the highest accuracy, and the accuracy of the baseline model is the worst.

\begin{table}[htbp]
\centering  
\caption{\textbf{Performance Comparison between DAgger models and the KIDA model}. KIDA suffers from an accuracy drop but has a lower loss}
\label{table:models}
\begin{tabular}{cccccc}
\toprule
Model   & Top 1 Acc.     & Top 3 Acc.     & Top 5 Acc.     & Loss          & Cum. Reward     \\ \midrule
Model 0 & \textbf{0.850} & \textbf{0.957} & \textbf{0.981} & 7.38          & 5304.9          \\
Model 1 & 0.797          & 0.917          & 0.966          & 9.50          & 5319.8          \\
Model 2 & 0.795          & 0.916          & 0.961          & 6.18          & 5237.5          \\
KIDA     & 0.721          & 0.822          & 0.870          & \textbf{2.96} & \textbf{7561.6} \\ \bottomrule
\end{tabular}
\end{table}

Models used in KIDA are the top $3$ performance models of DAgger. Table \ref{table:models} shows the comparison between these models and the KIDA model. Although KIDA has lower accuracy in collected validation data, it has a lower loss value and higher reward.

\subsection{Workload Apportionment}

In Workload Apportionment, baselines are trained and compared with random policy. From the table, we can see that the top $1$ accuracy of the baseline model is 45.6\%. The accuracy of a classification task with more than one hundred labels shows that the model has learned much expert knowledge. However, random strategies that do not rely on prior knowledge can obtain higher cumulative rewards.

\begin{table}[!htbp]
\centering  
\caption{\textbf{Performance Comparison between Baseline and Random Policy}. The baseline model gets great accuracy but fails to achieve higher rewards}
\begin{tabular}{ccccc}
\toprule
    & Top 1 Acc.     & Top 3 Acc.     & Top 5 Acc. & Cum. Reward     \\ \midrule
Baseline & \textbf{0.456} & \textbf{0.729} & \textbf{0.820}           & 624043.6          \\
Random & 0.013          & 0.034          & 0.052                    & \textbf{624928.9}          \\ \bottomrule
\end{tabular}
\end{table}

\section{Discussion}
\label{discuss}
Overall, we show that Strong Branching can not produce completely reliable labels. To further explore the performance of Strong Branching, we compare the variation of the dual bound using different methods.

\begin{figure}[h]
\centering  
\subfigure[Item Placement]{  
\begin{minipage}{0.33\linewidth}
\centering    
\includegraphics[scale=0.3]{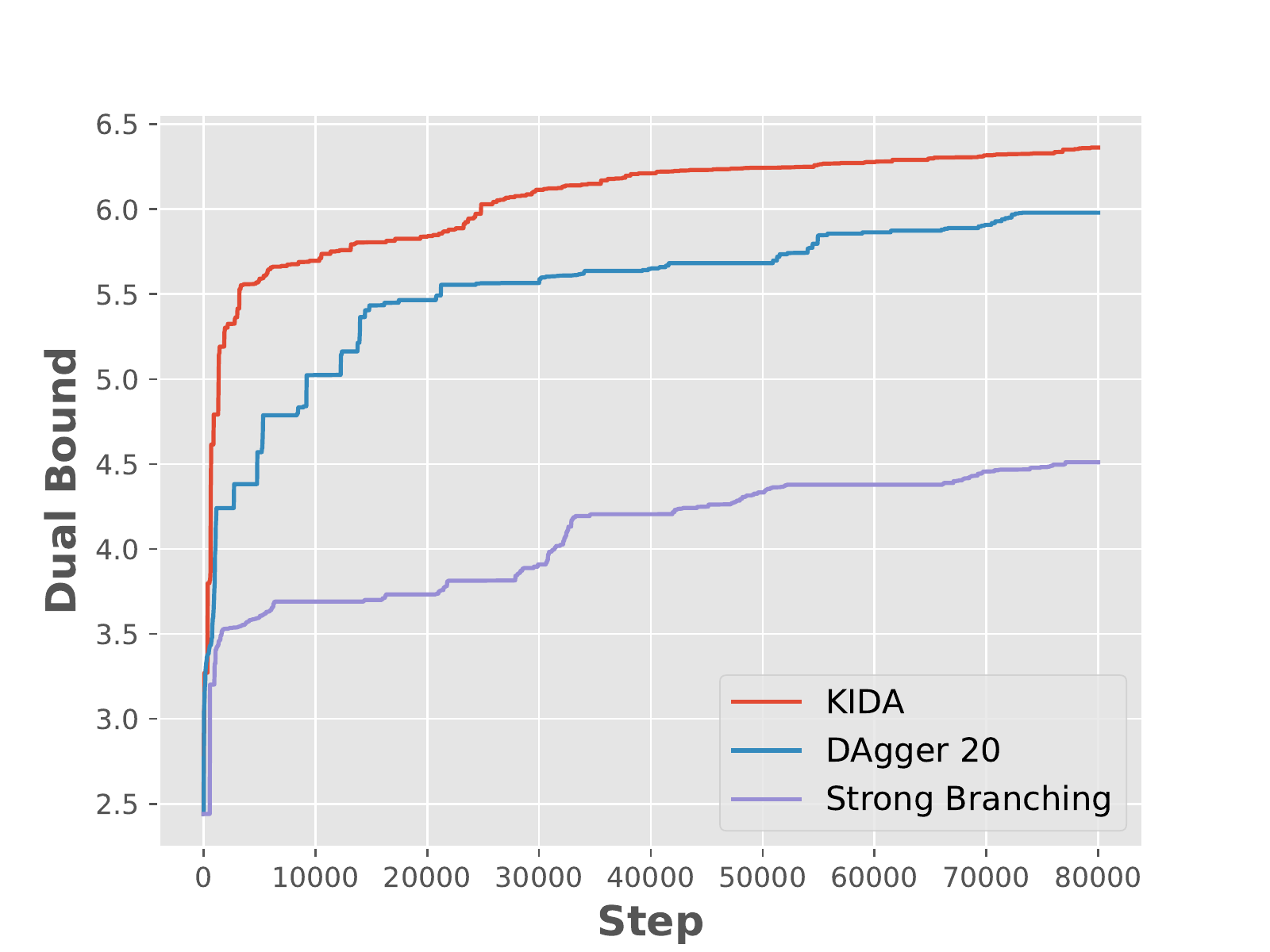}  
\end{minipage}
}\subfigure[Load Balancing]{
\begin{minipage}{0.33\linewidth}
\centering    
\includegraphics[scale=0.3]{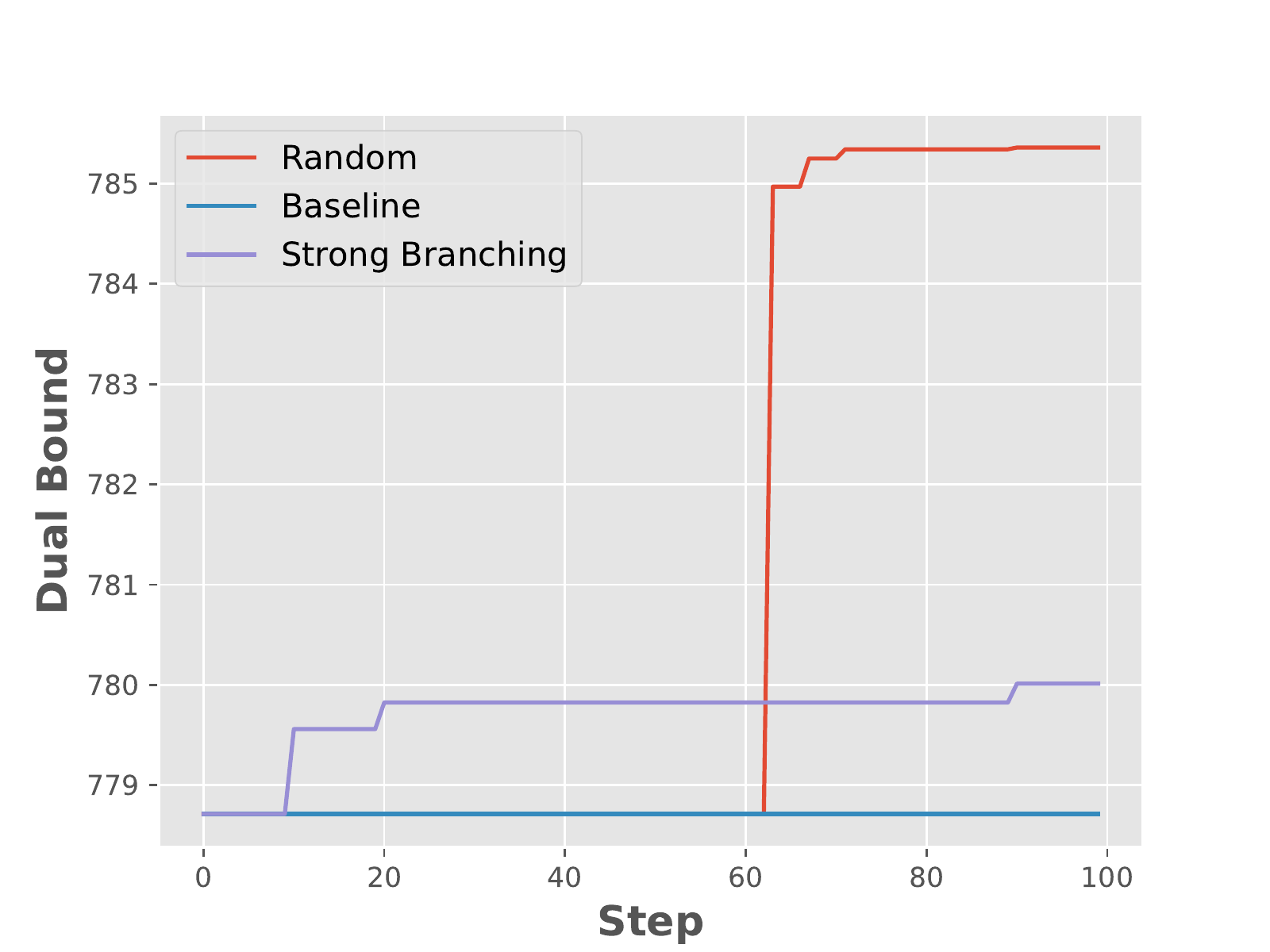}
\end{minipage}
}\subfigure[Anonymous]{
\begin{minipage}{0.33\linewidth}
\centering    
\includegraphics[scale=0.3]{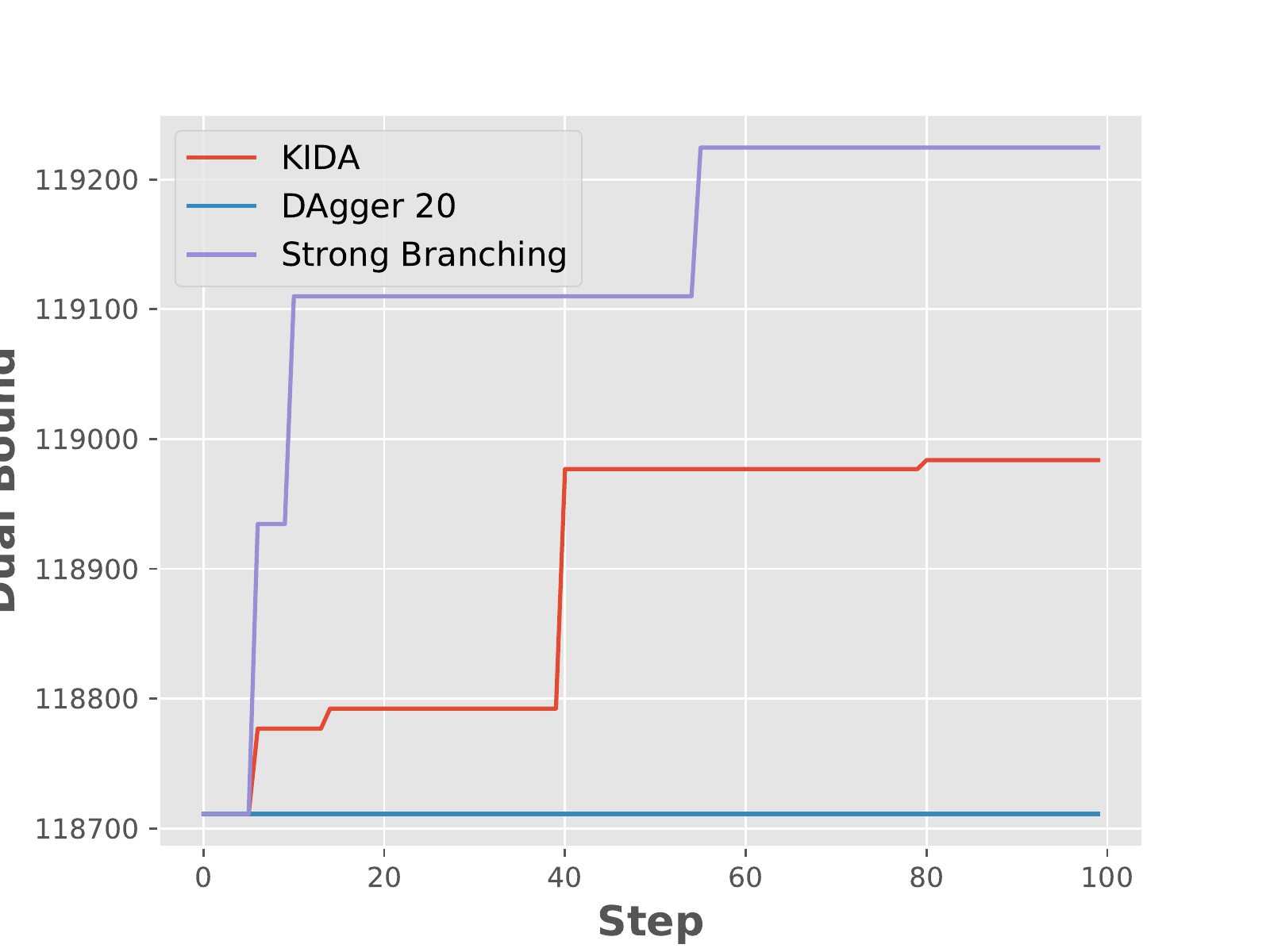}
\end{minipage}
}
\caption{\textbf{Dual Bound Improvements with Steps}. Strong Branching may still fall behind GNN-based policy or even random policy}   
\label{fig:dual_bound}    
\end{figure}

As shown in Figure~\ref{fig:dual_bound}, we select a random instance for each benchmark, the variation of the dual bound using different strategies. Even if we ignore the expensive time cost of Strong Branching, the dual bound improvement may still fall behind GNN models. This observation may be explained by dual degeneracy~\cite{greenberg1986analysis}. If there is a high dual degeneracy in LP solution, the product score of Strong Branching that SCIP uses to combine the improvements of the two-child nodes would be close to zero~\cite{gamrath2020scip}. In that case, Strong Branching may fail, and the expert knowledge needs to be redesigned based on other information accumulated in the problem-solving process.

\section{Conclusion}
\label{conclude}
In this paper, we devise a knowledge inheritance method in the dataset aggregation process for the dual task of ML4CO competition. By dataset aggregation, the inconsistency between training and deployment is reduced. By averaging the weights of different models from the dataset aggregation process, learned knowledge is generalized to get better results. Our model gets great improvements on Item Placement and Anonymous. Further, our experiments show models closer to expert knowledge do not necessarily achieve better results, which indicates Strong Branching fails in some cases. More heuristic branching algorithms need to be taken into account to build more reasonable expert knowledge to facilitate learning to branch.

{\small
\bibliographystyle{ieee_fullname}
\bibliography{main}
}

\end{document}